\documentclass{article}

\usepackage[preprint]{neurips_2023}

\usepackage[utf8]{inputenc} 
\usepackage[T1]{fontenc}    
\usepackage{hyperref}       
\usepackage{url}            
\usepackage{booktabs}       
\usepackage{amsfonts}       
\usepackage{nicefrac}       
\usepackage{microtype}      
\usepackage{xcolor}         

\usepackage{caption}
\usepackage{subcaption}
\usepackage{amsmath}
\usepackage{graphicx}

\usepackage[nolist]{acronym}

\def\nikhil#1{} 

\title{Neural Bee Colony Optimization: A Case Study in Public Transit Network Design}

\author{%
  Andrew Holliday \\
  School of Computer Science\\
  McGill University\\
  Montreal, QC \\
  \texttt{ahollid@cim.mcgill.ca} \\
  \And
  Gregory Dudek \\
  School of Computer Science \\
  McGill University \\
  \texttt{dudek@cim.mcgill.ca} \\
}

\begin{document}

\maketitle

\begin{abstract}
  In this work we explore the combination of metaheuristics and learned neural network solvers for combinatorial optimization.  We do this in the context of the \textbf{transit network design problem}, a uniquely challenging combinatorial optimization problem with real-world importance.  We train a neural network policy to perform single-shot planning of individual transit routes, and then incorporate it as one of several sub-heuristics in a modified Bee Colony Optimization (BCO) metaheuristic algorithm.  Our experimental results demonstrate that this hybrid algorithm outperforms the learned policy alone by up to 20\% and the original BCO algorithm by up to 53\% on realistic problem instances.  We perform a set of ablations to study the impact of each component of the modified algorithm.

\end{abstract}

\section{Introduction}




The design of urban transit networks is an important real-world problem, but is computationally very challenging.  It has some similarities with other \ac{CO} problems such as the \ac{TSP} and \ac{VRP}, but due to its many-to-many nature, combined with the fact that demand can be satisfied by transfers between transit lines, the problem is much more complex than those  well-studied problems.  The most successful approaches to the~\ac{NDP} to-date have been metaheuristic algorithms.  Metaheuristics are high-level approximate strategies for problem-solving that are agnostic to the kind of problem.  Many are inspired by natural phenomena, such as \ac{SA}, \ac{GA}, and \ac{BCO}.  

Metaheuristic algorithms have proven useful and remain the state-of-the-art in several very complex optimization problems~\citep{ahmed2019hyperheuristic}.  But little cross-over exists between the literature on this problem and that of machine learning with neural networks.  In this work, we use a neural network system to learn low-level heuristics for the \ac{NDP}, and use these learned heuristics in a metaheuristic algorithm.  We show that this synthesis of a machine learning approach and meta-heuristic approach outperforms either of them alone.

We first develop a novel \ac{GNN} policy model and train it in an \ac{RL} context to output transit networks that minimize an
established cost function.  We compare the performance of the trained \ac{GNN} model to that of \citet{nikolic2013transit}'s \ac{BCO} approach on a standard benchmark of \ac{NDP} instances~\citep{mumford2013dataset}, characterizing them over a range of different cost functions.  We then integrate this model into a metaheuristic algorithm called \ac{BCO}, as one of the heuristics that the algorithm can employ as it performs a stochastic search of the solution space.  We compare this approach to the \ac{GNN} model and the unmodified \ac{BCO} algorithm, and we find that on realistically-sized problem instances, the combination outperforms the \ac{GNN} by up to 20\% and \ac{BCO} by up to 53\%.  Lastly, we perform several ablations to understand the importance of different components of the proposed system to its performance.

\section{Related Work}\label{sec:relatedWorks}

\subsection{Graph Networks and Reinforcement Learning for Optimization Problems}

\acfp{GNN} are neural network models that are designed to operate on graph-structured data \citep{bruna2013spectral, kipf2016semi, defferrard2016spectral, duvenaud2015convolutional}. They were inspired by the success of convolutional neural nets on computer vision tasks and have been applied in many domains, including analyzing large web graphs~\citep{ying2018webscale}, designing printed circuit boards~\citep{mirhoseini2021graph}, and predicting chemical properties of molecules~\citep{duvenaud2015convolutional, gilmer2017quantum}. An overview of \acp{GNN} is provided by~\cite{battaglia2018relational}.

There has recently been growing interest in the application of machine learning techniques to solve \ac{CO} problems such as the \ac{TSP} and \ac{VRP}~\citep{bengio2021machine}.  As many such problems have natural interpretations as graphs, a popular approach has been to use \acp{GNN} to solve them.  A prominent early example is the work of~\citet{vinyals2015pointer}, who propose Pointer Networks and train them via supervised learning to solve \ac{TSP} instances.  

In \ac{CO} problems generally, it is difficult to find a globally optimal solution but easier to compute a scalar quality metric for any given solution.  As noted by~\citet{bengio2021machine}, this makes \ac{RL}, in which a system learns to maximize a scalar reward, a natural fit.  Recent work~\citep{dai2017learningCombinatorial, Kool2019AttentionLT, lu2019learning, sykora2020multi} has used \ac{RL} to train \ac{GNN} models and have attained impressive performance on the \ac{TSP}, the \ac{VRP}, and related problems. 

The solutions from some neural methods come close to the quality of those from specialized ~\ac{TSP} algorithms such as Concorde~\citep{concordeTspSolver}, while requiring much less run-time to compute~\citep{Kool2019AttentionLT}.  However, these methods all learn heuristics for constructing a single solution to a single problem instance.  By the nature of NP-hard problems such heuristics will always be limited in the quality of their results; In this work, we show that a metaheuristic algorithm that searches over multiple solutions from the learned heuristic can offer better quality.

\subsection{Optimization of Public Transit}

The transportation optimization literature has extensively studied the \acl{NDP}.  This problem is NP-complete~\citep{quak2003bus}, making it impractical to find optimal solutions for most cases.  While analytical optimization and mathematical programming methods have been successful on small instances~\citep{vannes2003AnalyticRouteAndSchedule, guan2006AnalyticRoutePlanning}, they struggle to realistically represent the problem~\citep{guihaire2008transitReview, kepaptsoglou2009transitReview}, and so metaheuristic approaches (as defined by~\citet{sorensen2018history}) have been more widely applied.  Historically, \acp{GA}, \ac{SA}, and ant-colony optimization have been most popular, along with hybrids of these methods~\citep{guihaire2008transitReview, kepaptsoglou2009transitReview}. But more recent work has adapted other metaheuristic algorithms such as \ac{BCO}~\citep{nikolic2013transit} and sequence-based selection hyper-heuristics~\citep{ahmed2019hyperheuristic}, demonstrating that they outperform approaches based on \acp{GA} and \ac{SA}.

On the other hand, while much work has used neural networks for predictive problems in urban mobility~\citep{xiong1992transportation, rodrigueNNsForLandUseAndTransport, chien2002dynamic, jeong2004bus, akgungorNNsForAccidentPrediction, li2020graph} and for other transit optimization problems such as scheduling and passenger flow control~\citep{zou2006lightrail, ai2022deep, Yan2023DistributedMD, jiang2018passengerInflow}, relatively little work has applied \ac{RL} or \acp{NN} to the \ac{NDP}. ~\citet{darwish2020optimising} and ~\citet{yoo2023reinforcement} both use \ac{RL} to design a network and schedule for the Mandl benchmark~\citep{mandl1980evaluation}, a single small graph with just 15 nodes. ~\citet{darwish2020optimising} use a \ac{GNN} approach inspired by~\cite{Kool2019AttentionLT}, while ~\citet{yoo2023reinforcement} uses tabular \ac{RL}.  Tabular \ac{RL} approaches tend to scale poorly; meanwhile, in our own work we experimented with a nearly identical approach to~\citet{darwish2020optimising}, but found it did not scale beyond very small instances.  Both these approaches also require a new model to be trained on each problem instance.  The technique developed here, by contrast, is able to find good solutions for realistically-sized \ac{NDP} instances of more than 100 nodes, and can be applied to problem instances unseen during training.

\section{The Transit Network Design Problem}

In the \ac{NDP}, one is given an augmented city graph $\mathcal{C} = (\mathcal{N}, \mathcal{E}_s, D)$, comprised of a set of $n$ nodes $\mathcal{N}$ representing candidate stop locations; a set of street edges $(i, j, \tau_{ij})$ connecting the nodes, with weights $\tau_{ij}$ indicating drive times on those streets; and an $n \times n$ \ac{OD} matrix $D$ giving the travel demand (in number of trips) between every pair of nodes in $\mathcal{N}$.  The goal is to propose a set of routes $\mathcal{R}$, where each route $r$ is a sequence of nodes in $\mathcal{N}$, so as to minimize a cost function $C: \mathcal{C}, \mathcal{R} \rightarrow \mathbb{R}^+$.  $\mathcal{R}$ is also subject to the following constraints:
\begin{enumerate}
    \item The route network $\mathcal{R}$ must be connected, allowing every node in $\mathcal{N}$ to be reached from every other node via transit.
    \item The route network must contain exactly $S$ routes, that is, $|\mathcal{R}| = S$, where $S$ is a parameter set by the user.
    \item Every route $r \in \mathcal{R}$ must be within stop limits $MIN \leq |r| \leq MAX$, where $MIN$ and $MAX$ are parameters set by the user.
    \item No route $r \in \mathcal{R}$ may contain cycles; that is, it must include each node $i$ at most once.
\end{enumerate}
We here deal with the symmetric \ac{NDP}, that is: $D = D^\top$,
$(i, j, \tau_{ij}) \in \mathcal{E}_s$ iff. $(j, i, \tau_{ij}) \in \mathcal{E}_s$, and all routes are traversed both forwards and backwards by vehicles on them.

\subsection{Markov Decision Process Formulation}\label{subsec:mdp}

A \acf{MDP} is a formalism for describing a step-by-step problem-solving process, commonly used to define problems in \ac{RL}.  In an \ac{MDP}, an \textbf{agent} interacts with an environment over a series of time steps.  At each time step $t$, the environment is in some \textbf{state} $s_t \in \mathcal{S}$; the agent takes some \textbf{action} $a_t$ which belongs to the set $\mathcal{A}_t$ of available actions in state $s_t$.  This causes a transition to a new state $s_{t+1} \in \mathcal{S}$ according to the state transition distribution $P(s' | s, a)$, and also gives the agent a numerical \textbf{reward} $R_t \in \mathbb{R}$ according to the reward distribution $P(R | s, a, s')$.  The agent acts according to a \textbf{policy} $\pi(a|s)$, which is a probability distribution over the available actions in each state.  In \ac{RL}, the goal is typically to \textbf{learn} a policy $\pi$ through repeated interactions with the environment, such that $\pi$ maximizes some measure of reward over time.

We here describe the \ac{MDP} we use to represent the \acl{NDP}.  At a high level, the \ac{MDP} alternates at every step $t$ between two modes: \textbf{extend}, where the agent selects an extension to the route $r_t$ that it is currently planning; and \textbf{halt}, where the agent chooses whether to continue extending $r_t$ or stop, adding it as-is to the transit network and beginning the planning of a new route.  This alternation is captured by the state variable $\textbf{extend}_t \in \{\textup{True}, \textup{False}\}$, a boolean which changes its value after every step:
\begin{align}
\textbf{extend}_t = \begin{cases} \neg \textbf{extend}_{t-1} & \text{if $t > 0$} \\
\textup{False} & \text{otherwise} \end{cases}
\end{align}
More completely, the state $s_t$ is composed of the city graph $\mathcal{C}$, the set of routes $\mathcal{R}_t$ planned so far, the state of the in-progress route $r_t$, and the mode variable $\textbf{extend}_t$.  As $\mathcal{C}$ does not change with $t$, we represent the $s_t$ as in eqn.~\ref{eqn:state}.
\begin{equation}\label{eqn:state}
    s_t = (\mathcal{R}_t, r_t, \textbf{extend}_t)
\end{equation}
The starting state is $s_0 = (\mathcal{R}_0 = \{\}, r_0 = [], \textbf{extend}_0 = \textup{True})$.

When the expression ($\textbf{extend}_t=\textup{True}$), the available actions are drawn from $\textup{SP}$, the set of shortest paths between all node pairs.  If $r_t = []$, then $\mathcal{A}_t = \{a | a \in \textup{SP}, |a| \leq MAX\}$.  Otherwise, $\mathcal{A}_t$ is comprised of paths $a \in \textup{SP}$ satisfying all of the following conditions:
\begin{itemize}
    \item $(i,j,\tau_{ij}) \in \mathcal{E}_s$, where $i$ is the first node of $a$ and $j$ is the last node of $r_t$, or vice-versa
    \item $a$ and $r_t$ have no nodes in common
    \item $|a| \leq MAX - |r_t|$
\end{itemize}
Once a path $a_t \in \mathcal{A}_t$ is chosen, $r_{t+1}$ is formed by appending $a_t$ to the beginning or end of $r_t$ as appropriate: $r_{t+1} = \textup{combine}(r_t, a_t)$.

When ($\textbf{extend}_t=\textup{False}$), the action space is given by eqn.~\ref{eqn:halt_actions}.
\begin{align}\label{eqn:halt_actions}
    \mathcal{A}_t = \begin{cases}
        \{\textup{continue}\} & \text{if} |r| < MIN \\
        \{\textup{halt}\} & \text{if} |r| = MAX \\        
        \{\textup{continue}, \textup{halt}\} & \text{otherwise}
    \end{cases}
\end{align}
If $a_t = \textup{halt}$, $r_t$ is added to $\mathcal{R}_t$ to get $\mathcal{R}_{t+1}$, and $r_{t+1} = []$ is a new empty route; if $a_t = \textup{continue}$, then $\mathcal{R}_{t+1}$ and $r_{t+1}$ are unchanged from step $t$.

Thus, the full state transition distribution is deterministic, and is described by eqn.~\ref{eqn:state_trans}.
\begin{align}\label{eqn:state_trans}
s_t = \begin{cases} 
    (\mathcal{R}_t = \mathcal{R}_{t-1}, r_t = \textup{combine}(r_{t-1}, a_{t-1}), \textup{False}) & \text{if $\textbf{extend}_{t-1}$} \\
    (\mathcal{R}_t = \mathcal{R}_{t-1} \cup \{r_{t-1}\}, r_t = [], \textup{True}) & \text{if $\neg \textbf{extend}_{t-1}$ and $a_{t-1} = \text{halt}$} \\
    (\mathcal{R}_t = \mathcal{R}_{t-1}, r_t = r_{t-1}, \textup{True}) & \text{if $\neg \textbf{extend}_{t-1}$ and $a_{t-1} = \text{continue}$}
\end{cases}
\end{align}

When $|\mathcal{R}_t| = S$, the \ac{MDP} terminates, giving the final reward $R_t = -C(\mathcal{C}, \mathcal{R}_t)$. The reward $R_t = 0$ at all prior steps.

This \ac{MDP} formalization imposes some helpful biases on the solution space.  First, it requires all transit routes to follow the street graph $(\mathcal{N}, \mathcal{E}_s)$; any route connecting $i$ and $j$ must also stop at all nodes along some path between $i$ and $j$, thus biasing planned routes towards covering more nodes.  Second, it biases routes towards being direct and efficient by forcing them to be composed of shortest paths; though in the limiting case a policy may construct arbitrarily indirect routes by choosing paths with length 2 at every step, this is unlikely as the majority of paths in $\textup{SP}$ are longer than two edges in realistic street graphs.  Thirdly and finally, the alternation between deciding to whether to continue a route and deciding to how to extend it means that the probability of halting does not depend on how many different extensions are possible.  

\subsection{Cost Function}

We can define the cost function in general as being composed of three components.  The cost to riders is the average time of all passenger trips over the network:
\begin{equation}
    C_p(\mathcal{C}, \mathcal{R}) = \frac{\sum_{i,j} D_{ij}\tau_{\mathcal{R}ij}}{\sum_{i,j} D_{ij}}
\end{equation}
Where $\tau_{\mathcal{R}ij}$ is the time of the shortest transit trip from $i$ to $j$ given $\mathcal{R}$, including a time penalty $p_T$ for each transfer.  The operating cost is the total driving time of the routes:
\begin{equation}
    C_o(\mathcal{C}, \mathcal{R}) = \sum_{r \in \mathcal{R}} \tau_r
\end{equation}
Where $\tau_r$ is the time needed to completely traverse a route $r$ in both directions.

To enforce the constraints on $\mathcal{R}$, we also add a term $C_c$, which is the fraction of node pairs that are not connected by $\mathcal{R}$ plus a measure of how much $|r| > MAX$ or $|r| < MIN$ across all routes.  The cost function is then:

\begin{equation}
    C(\mathcal{C}, \mathcal{R}) = \alpha w_p C_p(\mathcal{C}, \mathcal{R}) + (1 - \alpha) w_o C_o(\mathcal{C}, \mathcal{R}) + \beta C_c(\mathcal{C}, \mathcal{R})
\end{equation}

The weight $\alpha \in [0, 1]$ controls the trade-off between passenger and operator costs.  $w_p$ and $w_o$ are re-scaling constants chosen so that $w_p C_p$ and $w_o C_o$ both vary roughly over the range $[0, 1)$ for different $\mathcal{C}$ and $\mathcal{R}$; this is done so that $\alpha$ will properly balance the two, and to stabilize training of the \acl{NN} policy.  The values used are $w_p = (\max_{i,j}T_{ij})^{-1}$ and $w_o = (3S\max_{i,j}T_{ij})^{-1}$, where $T$ is an $n \times n$ matrix of shortest-path driving times between every node pair.

\section{Learned Planner}

We propose to learn a policy $\pi_\theta(a|s)$ with parameters $\theta$ with the objective of maximizing $G = \sum_t R_t$ on the \ac{MDP} described in section~\ref{subsec:mdp}.  Since reward is only given at the final timestep, we have: 
\begin{equation}
    G = -C(\mathcal{C}, \mathcal{R}_{final})
\end{equation}
By rolling out this policy on the \ac{MDP} with some city $\mathcal{C}$, we can obtain a transit network $\mathcal{R}$ for that city.  We denote this algorithm the Learned Planner (LP), or $\textup{LP}(\mathcal{C}, \alpha, \mathcal{R}_0 = \{\})$.

The policy $\pi_\theta$ is a neural network model parameterized by $\theta$.  Its ``backbone'' is a graph attention network \citep{gatv2conv} which treats the city as a fully-connected graph on the nodes $\mathcal{N}$, where each edge has an associated feature vector $\mathcal{e}_{ij}$ containing information about demand, existing transit connections, and the street edge (if one exists) between $i$ and $j$.  We note that a graph attention network operating on a fully-connected graph has close parallels to a Transformer model~\citep{vaswani2017attention}, but unlike Transformers this architecture enables the use of edge features that describe known relationships between elements.

The backbone \ac{GNN} outputs node embeddings $Y$, which are operated on by one of two policy ``heads'', depending on the state $s_t$: $\textup{NN}_{ext}$ for choosing among extensions when $\textbf{extend}_t=\textup{True}$, and $\textup{NN}_{halt}$  for deciding whether to halt when $\textbf{extend}_t=\textup{False}$.  The details of the network architecture are provided in Appendix B in the supplementary material.

\subsection{Training}\label{subsec:methodology_training}

Following the work of~\citet{Kool2019AttentionLT}, we train the policy network using the policy gradient method REINFORCE with baseline~\citep{williams1992reinforce} and set $\gamma = 1$.  Since the reward $R_t$ for the last step is the negative cost and at all other steps $R_t=0$, by setting the discount rate $\gamma = 1$, the return $G_t$ to each action $a_t$ is simply $G_t = \sum_{t'} \gamma^{t' - t} R_t = -C(\mathcal{C}, \mathcal{R})$.  The learning signal for each action $a_t$ is $G_t - \textup{baseline}(\mathcal{C}, \alpha)$, where the baseline function $\textup{baseline}(\mathcal{C}, \alpha)$ is a separate \ac{MLP} trained to predict the final reward obtained by the current policy for a given cost weight $\alpha$ and city $\mathcal{C}$.

The model is trained on a variety of synthetic cities and over a range of values of $\alpha \in [0, 1]$.  $S, n, MIN,$ and $MAX$ are held constant during training.  For each batch, a full rollout of the MDP episode is performed, the cost is computed, and back-propagation and weight updates are applied to both the policy network and the baseline network.

Each synthetic city begins construction by generating its nodes and street network using one of these processes chosen at random:
\begin{itemize}
    \item Incoming $4$-nn: Sample $n$ random 2D points uniformly in a square to give $\mathcal{N}$.  Add street edges to each node $i$ from it's four nearest neighbours.
    \item Outgoing $4$-nn: The same as the above, but add edges in the opposite direction.
    \item Voronoi: Sample $m$ random 2D points, and compute their Voronoi diagram~\citep{fortune1995voronoi}.  Take the shared vertices and edges of the resulting Voronoi cells as $\mathcal{N}$ and $\mathcal{E}_s$.  $m$ is chosen so $|\mathcal{N}| = n$.
    \item 4-grid: Place $n$ nodes in a rectangular grid as close to square as possible.  Add edges from each node to its horizontal and vertical neighbours.
    \item 8-grid: The same as the above, but also add edges between diagonal neighbours.
\end{itemize}

For all models except Voronoi, each edge is then deleted with user-defined probability $\rho$.  If the resulting street graph is not strongly connected (that is, all nodes are reachable from all other nodes), it is discarded and the process is repeated.  Nodes are sampled in a $30 \textup{km} \times 30 \textup{km}$ square, and a fixed vehicle speed of $v = 15 \textup{m/s}$ is assumed to compute street edge weights $\tau_{ij} = ||(x_i, y_i) - (x_j, y_j)||_2 / v$.  Finally, we generate the \ac{OD} matrix $D$ by setting diagonal demands $D_{ii} = 0$ and uniformly sampling off-diagonal elements $D_{ij} \sim [60, 800]$.


All neural network inputs are normalized so as to have unit variance and zero mean across the entire dataset during training.  The scaling and shifting normalization parameters are saved as part of the model and applied to new data presented at test time. 

\section{Bee Colony Optimization}

\ac{BCO} is an algorithm inspired by how bees in a hive cooperate to search for nectar.  At a high level, it works as follows.  Given an initial problem solution $\mathcal{R}_0$ and a cost function $C$, a fixed number $B$ of ``bee'' processes are initialized with $\mathcal{R}_b = \mathcal{R}_0 \; \forall \; b \in [0,B]$.  Each bee makes a fixed number $N_C$ of random modifications to $\mathcal{R}_b$, discarding the modification if it increases cost $C(\mathcal{R}_b)$.  Then each bee is randomly designated a ``recruiter'' or ``follower'', where $P(b = \textup{follower}) \propto C(\mathcal{R}_b)$.  Each follower bee $b_f$ copies the solution of a random recruiter bee $b_r$, with probability inversely related to $C(\mathcal{R}_{b_r})$.  These alternating steps of exploration and recruitment are repeated until some termination condition is met, and the lowest-cost solution $\mathcal{R}_{best}$ found over the process is returned. 
 
In~\citet{nikolic2013transit}, \ac{BCO} is adapted to the \ac{NDP} by dividing the worker bees into two types, which apply different random modification processes.  Given a network $\mathcal{R}_b$ with $S$ routes for city $\mathcal{C}$ for each bee $b$, each bee selects a route $r \in \mathcal{R}_b$ with probability inversely related to the amount of demand $r$ directly satisfies, and then selects a random terminal (first or last node) on $r$.  Type-1 bees replace the chosen terminal with a random other terminal node in $\mathcal{N}$, and make the new route the shortest path between the new terminals.  Meanwhile, type-2 bees choose with probability $0.2$ to delete the chosen terminal from the route, and with probability $0.8$ to add a random node neighbouring the chosen terminal to the route (at the start or end, depending on the terminal), making that node the new terminal.  The overall best solution is updated after every $N_P$ modification-and-recruitment steps (making one ``iteration''), and the algorithm performs $I$ iterations before halting.  Henceforth, ``\ac{BCO}'' refers specifically to this \ac{NDP} algorithm.

We propose a modification of this algorithm, called Neural \ac{BCO} (``NBCO'' henceforth), in which the type-1 bees are replaced by ``neural bees''.  A neural bee selects a route $r \in \mathcal{R}$ for modification in the same manner as the type-1 and type-2 bees, but instead of selecting a terminal on $r$, it rolls out our learned policy $\pi_\theta$ to replace $r$ with a new route $r' \leftarrow \textup{LP}(\mathcal{C}, \alpha, \mathcal{R} \setminus \{r\})$.  We replace the type-1 bee because its action space (replacing one route by a shortest path) is a subset of the action space of the neural bee (replacing one route by a new route composed of shortest paths), while the type-2 bee's action space is quite different.  The algorithm is otherwise unchanged; for the full details, we refer the reader to~\citet{nikolic2013transit}.

\section{Experiments}

In all experiments, the policies $\pi_\theta$ used are trained on a dataset of $2^{15} = 32,768$ synthetic cities with $n = 20$.  A 90:10 training:validation split of this dataset is used; after each epoch of training, the model is evaluated on the validation set, and at the end of training, the model weights from the epoch with the best validation-set performance are returned.  Data augmentation is applied each time a city is used for training.  This consists of multiplying the node positions $(x_i, y_i)$ and travel times $\tau_{ij}$ by a random factor $c_s \sim [0.4, 1.6]$, rotating the node positions about their centroid by a random angle $\phi \sim [0^\circ, 360^\circ)$, and multiplying $D$ by a random factor $c_d \in [0.8, 1.2]$.  During training and evaluation, constant values $S=10, MIN=2, MAX=15$ are used.  Training proceeds for 5 epochs, with a batch size of 64 cities.  When training with different random seeds, the dataset is held constant across seeds but the data augmentation is not.

All evaluations are performed on the Mandl~\citep{mandl1980evaluation} and Mumford~\citep{mumford2013dataset} city datasets, two popular benchmarks for evaluating \ac{NDP} algorithms~\citep{mumford2013new, john2014routing, kilic2014demand, ahmed2019hyperheuristic}.  The Mandl dataset is one small synthetic city, while the Mumford dataset consists of four synthetic cities, labelled Mumford0 through Mumford3, that range in size from $n=30$ to $n=127$, and gives values of $S, MIN,$ and $MAX$ to use when benchmarking on each city.  The values $n, S, MIN,$ and $MAX$ for Mumford1, Mumford2, and Mumford3 are taken from three different real-world cities and their existing transit networks, giving the dataset a degree of realism.  Details of these benchmarks are given in Table~\ref{tab:dataset}.

\begin{table}[t]
    \caption{Statistics of the five benchmark problems used in our experiments.}    
    \label{tab:dataset}
    \centering
    \begin{tabular}{lcccccc}
        \toprule
          City & \# nodes $n$ & \# street edges $|\mathcal{E}_s|$ & \# routes $S$ & $MIN$ & $MAX$ & Area (km$^2$) \\
          \midrule
          Mandl   & 15  & 20  & 6  & 2  & 8 & 352.7 \\
         Mumford0 & 30  & 90  & 12 & 2  & 15 & 354.2 \\
         Mumford1 & 70  & 210 & 15 & 10 & 30 & 858.5 \\
         Mumford2 & 110 & 385 & 56 & 10 & 22 & 1394.3 \\
         Mumford3 & 127 & 425 & 60 & 12 & 25 & 1703.2 \\
         \bottomrule
    \end{tabular}
\end{table}

For both \ac{BCO} and NBCO, we set all algorithmic parameters to the values used in the experiments of~\citet{nikolic2013transit}: $B=10, N_C=2, N_P=5, I=400$.  We also ran \ac{BCO} for up to $I=2,000$ on several cities, but found this did not yield any improvement over $I=400$.  We run NBCO with equal numbers of neural bees and type-2 bees, just as \ac{BCO} uses equal numbers of type-1 and type-2 bees.  Hyperparameter settings of the policy's model architecture and training process were arrived at by a limited manual search; for their values, we direct the reader to the configuration files contained in our code release.  We set the constraint penalty weight $\beta=5$ in all experiments.

\subsection{Results}

We compare LP, \ac{BCO}, and NBCO on Mandl and the four Mumford cities.  To evaluate LP, we perform 100 rollouts and choose the lowest-cost $\mathcal{R}$ from among them (denoted LP-100).  Each algorithm is run across a range of 10 random seeds, with a separate policy network trained with that seed.  We report results averaged over all of the seeds.  Our main results are summarized in Table~\ref{tab:main}, which shows results at three different $\alpha$ values, $0.0, 1.0,$ and $0.5$, which optimize for the operators' perspective, the passengers' perspective, and a balance of the two.  This table also contains results for two ablation experiments: one in which LP was rolled out $40,000$ times instead of 100 (denoted LP-40k), and one in which we ran a variant of NBCO with only neural bees, no type-2 bees.

The results show that while \ac{BCO} performs best on the two smallest cities in most cases, its relative performance worsens considerably when $n = 70$ or more.  On Mumford1, 2, and 3, for each $\alpha$, LP matches or outperforms \ac{BCO}.  Meanwhile, NBCO with a mixture of bee types performs best overall on these three cities.  It is better than LP-100 in every instance, improving on its cost by about 6\% in most cases at $\alpha=1.0$ and $0.5$, and by up to 20\% at $\alpha=0.0$; and it improves on \ac{BCO} by 33\% to 53\% on Mumford3 depending on $\alpha$.

NBCO does fail to obey route length limits on 1 out of 10 seeds when $\alpha=0.0$.  This may be due to $\alpha=0.0$ causing the benefits from under-length routes overwhelm the cost penalty due to a few routes being too long.  
This could likely be resolved by simply increasing $\beta$ or adjusting the specific form of $C_c$.

\begin{table}[h]
    \centering
    \caption{Average final cost $C(\mathcal{C}, \mathcal{R}, \alpha)$ achieved by each method over 10 random seeds, for three different settings of cost weight $\alpha$.  Bold indicates the best in each column.  Orange indicates that one seed's solution violated a constraint, red indicates two or three seeds did so.  Percentages are standard deviations over the 10 seeds.}
\begin{tabular}{cllllll}
\toprule
 City &             Mandl &          Mumford0 &          Mumford1 &          Mumford2 &          Mumford3 \\
 Method &        &           &           &           &           \\    
\midrule
 \multicolumn{6}{c}{$\alpha = 0.0$} \\
\midrule
 BCO &  \textbf{0.276 $\pm$ 17\%} &  \textbf{0.272 $\pm$ 16\%} &  0.854 $\pm$ 32\% &  0.692 $\pm$ 40\% &  \textcolor{orange}{0.853 $\pm$ 35\%} \\
 LP-100 &  0.317 $\pm$ 25\% &  0.487 $\pm$ 66\% &  0.853 $\pm$ 43\% &  0.688 $\pm$ 26\% &  0.710 $\pm$ 24\% \\
 LP-40k &  0.273 $\pm$ 26\% &  0.440 $\pm$ 68\% &  0.805 $\pm$ 41\% &  0.665 $\pm$ 27\% &  0.690 $\pm$ 25\% \\
 NBCO &  0.279 $\pm$ 20\% &  0.298 $\pm$ 41\% &  \textcolor{orange}{\textbf{0.623 $\pm$ 26\%}} &  \textcolor{orange}{\textbf{0.537 $\pm$ 44\%}} &  \textcolor{orange}{\textbf{0.572 $\pm$ 46\%}} \\
 No-2-NB &  0.290 $\pm$ 18\% &  0.295 $\pm$ 41\% &  0.670 $\pm$ 53\% &  0.605 $\pm$ 48\% &  0.574 $\pm$ 49\% \\    
\midrule
\multicolumn{6}{c}{$\alpha = 0.5$} \\
\midrule
BCO &   \textbf{0.328 $\pm$ 3\%} &   \textbf{0.563 $\pm$ 2\%} &  \textcolor{red}{1.015 $\pm$ 42\%} &  0.710 $\pm$ 36\% &  0.944 $\pm$ 29\% \\
LP-100 &  0.343 $\pm$ 10\% &  0.638 $\pm$ 22\% &  0.742 $\pm$ 24\% &  0.617 $\pm$ 14\% &  0.612 $\pm$ 13\% \\
LP-40k &   0.329 $\pm$ 9\% &  0.619 $\pm$ 20\% &  0.718 $\pm$ 22\% &  0.606 $\pm$ 14\% &  0.601 $\pm$ 14\% \\
NBCO &   0.331 $\pm$ 7\% &   0.571 $\pm$ 6\% &   \textbf{0.627 $\pm$ 8\%} &   \textbf{0.532 $\pm$ 5\%} &  \textbf{0.584 $\pm$ 37\%} \\
No-2-NB &   0.330 $\pm$ 4\% &   0.588 $\pm$ 7\% &  0.639 $\pm$ 15\% &  0.589 $\pm$ 34\% &  0.596 $\pm$ 37\% \\
\midrule
\multicolumn{6}{c}{$\alpha = 1.0$} \\
\midrule
BCO &   \textbf{0.314 $\pm$ 1\%} &   0.645 $\pm$ 5\% &  \textcolor{orange}{0.739 $\pm$ 37\%} &  0.656 $\pm$ 38\% &  \textcolor{red}{1.004 $\pm$ 40\%} \\
LP-100 &   0.335 $\pm$ 2\% &   0.738 $\pm$ 6\% &   0.600 $\pm$ 3\% &   0.534 $\pm$ 2\% &   0.504 $\pm$ 1\% \\
LP-40k &   0.325 $\pm$ 1\% &   0.709 $\pm$ 5\% &   0.587 $\pm$ 3\% &   0.528 $\pm$ 2\% &   0.498 $\pm$ 1\% \\
NBCO &   0.317 $\pm$ 1\% &   \textbf{0.637 $\pm$ 6\%} &   \textbf{0.564 $\pm$ 3\%} &   \textbf{0.507 $\pm$ 1\%} &   \textbf{0.481 $\pm$ 2\%} \\
No-2-NB &   0.320 $\pm$ 1\% &   0.668 $\pm$ 7\% &   0.570 $\pm$ 2\% &   0.511 $\pm$ 2\% &   0.486 $\pm$ 2\% \\
\bottomrule
\end{tabular}

    \label{tab:main}
\end{table}

\subsection{Ablations}

We first observe that under the parameter settings used here, \ac{BCO} and NBCO both consider a total of $B \times N_C \times N_P \times I = 40,000$ different networks over a single run.  To see whether NBCO's improvement over LP-100 is simply due to its exploring more solutions, we performed the LP-40k experiments, taking $40,000$ samples from LP instead of $100$.  The results for LP-40k in Table~\ref{tab:main} show that it while it improves on NBCO on Mandl, for all larger cities it is only slightly better than LP-100.  The gap between LP-40k and NBCO is at least 54\% larger than the gap between LP-40k and LP-100 on each Mumford city for each $\alpha$ value, and in 10 of the 12 cases it is more than twice as large.  This indicates that the main factor in NBCO's improvement over LP is the metaheuristic algorithm that guides the search through solution space.

To examine the importance of the type-2 bee to NBCO, we run another set of experiments with a variant of NBCO with no type-2 bees, only neural bees, denoted No-2-NB.  Again the results are displayed in Table~\ref{tab:main}.  We observe that with the exception of Mumford0 with $\alpha=0.0$ and Mandl with $\alpha=0.5$, its performance is worse than with both types of bees: the very different action space of the type-2 bees is a useful complement to the neural bees.  However, this variant still outperforms \ac{BCO} and both LP variants for most cities and $\alpha$ values: it is the guidance of the learned heuristic by the bee-colony metaheuristic that is responsible for most of NBCO's superior performance.

\subsection{Trade-offs Between Passenger and Operator Costs}

\begin{figure}[t]
    \centering
    \begin{subfigure}[b]{0.32\textwidth}
        \centering
        \includegraphics[width=\textwidth]{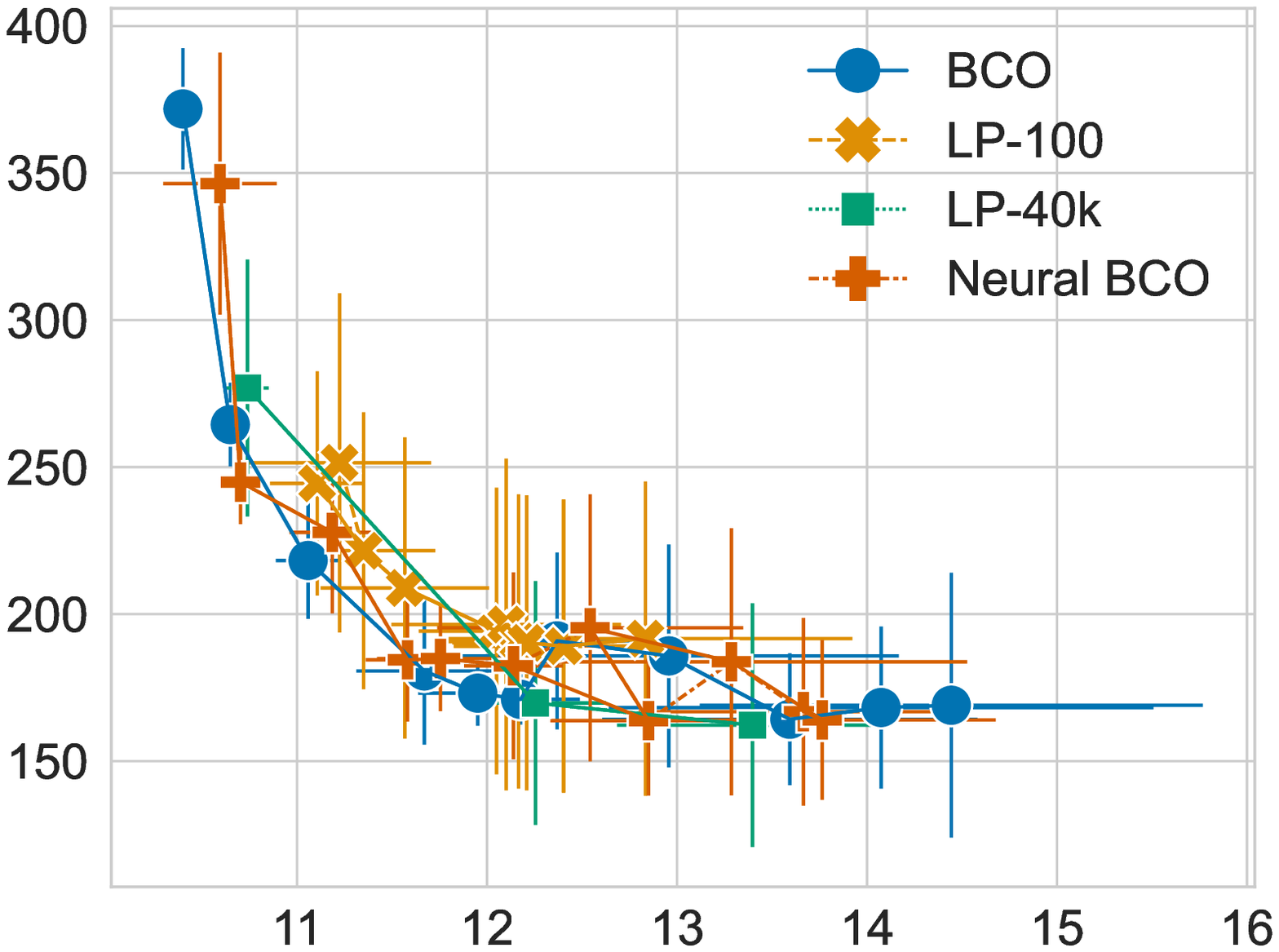}
        \caption{Mandl}
        \label{subfig:mandlpareto}
    \end{subfigure}
    \hfill
    \begin{subfigure}[b]{0.32\textwidth}
        \centering
        \includegraphics[width=\textwidth]{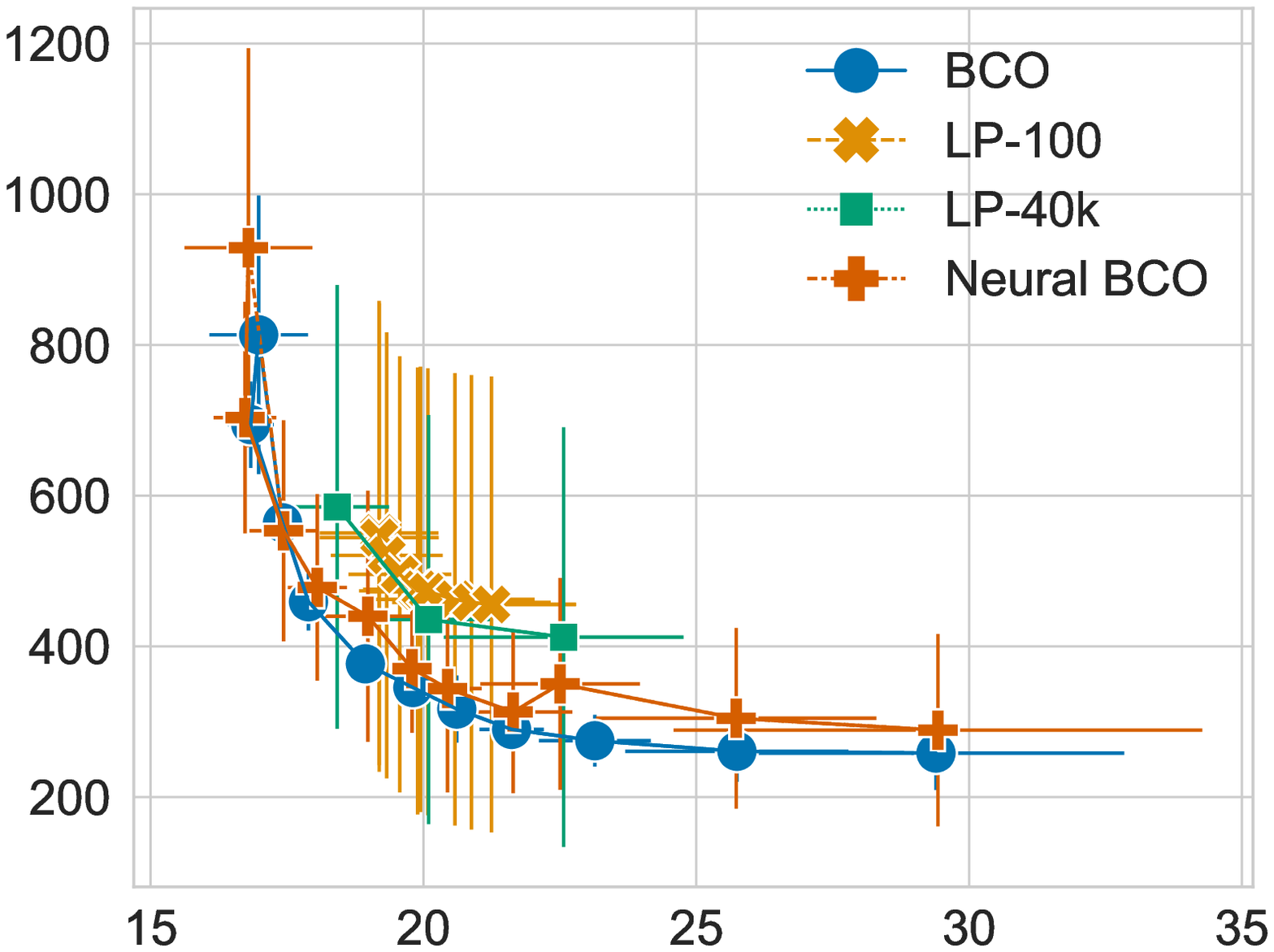}
        \caption{Mumford0}
        \label{subfig:mumford0pareto}
    \end{subfigure}
    \hfill
    \begin{subfigure}[b]{0.32\textwidth}
        \centering
        \includegraphics[width=\textwidth]{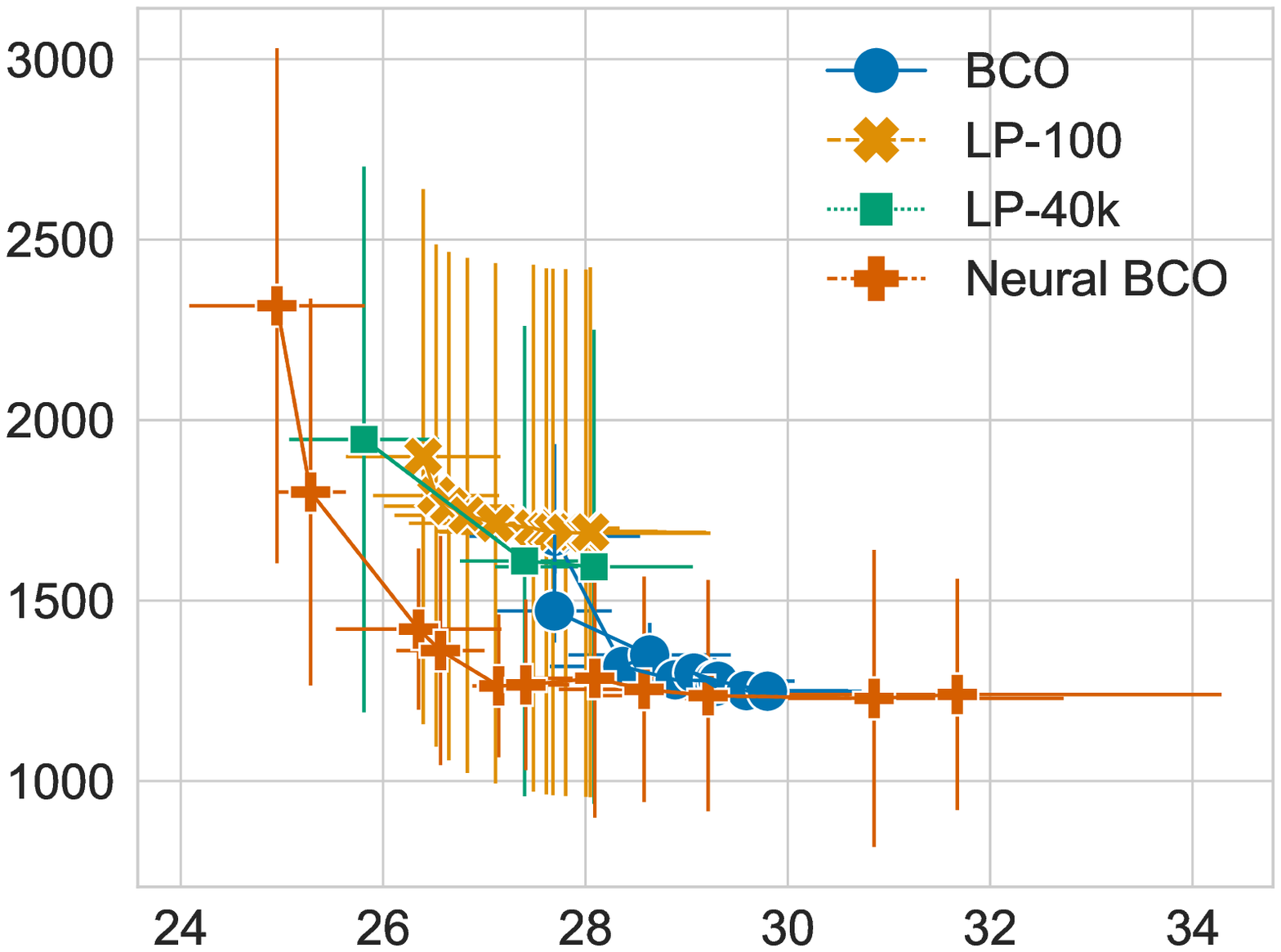}
        \caption{Mumford1}
        \label{subfig:mumford1pareto}
    \end{subfigure}
    \hfill
    \begin{subfigure}[b]{0.32\textwidth}
        \centering
        \includegraphics[width=\textwidth]{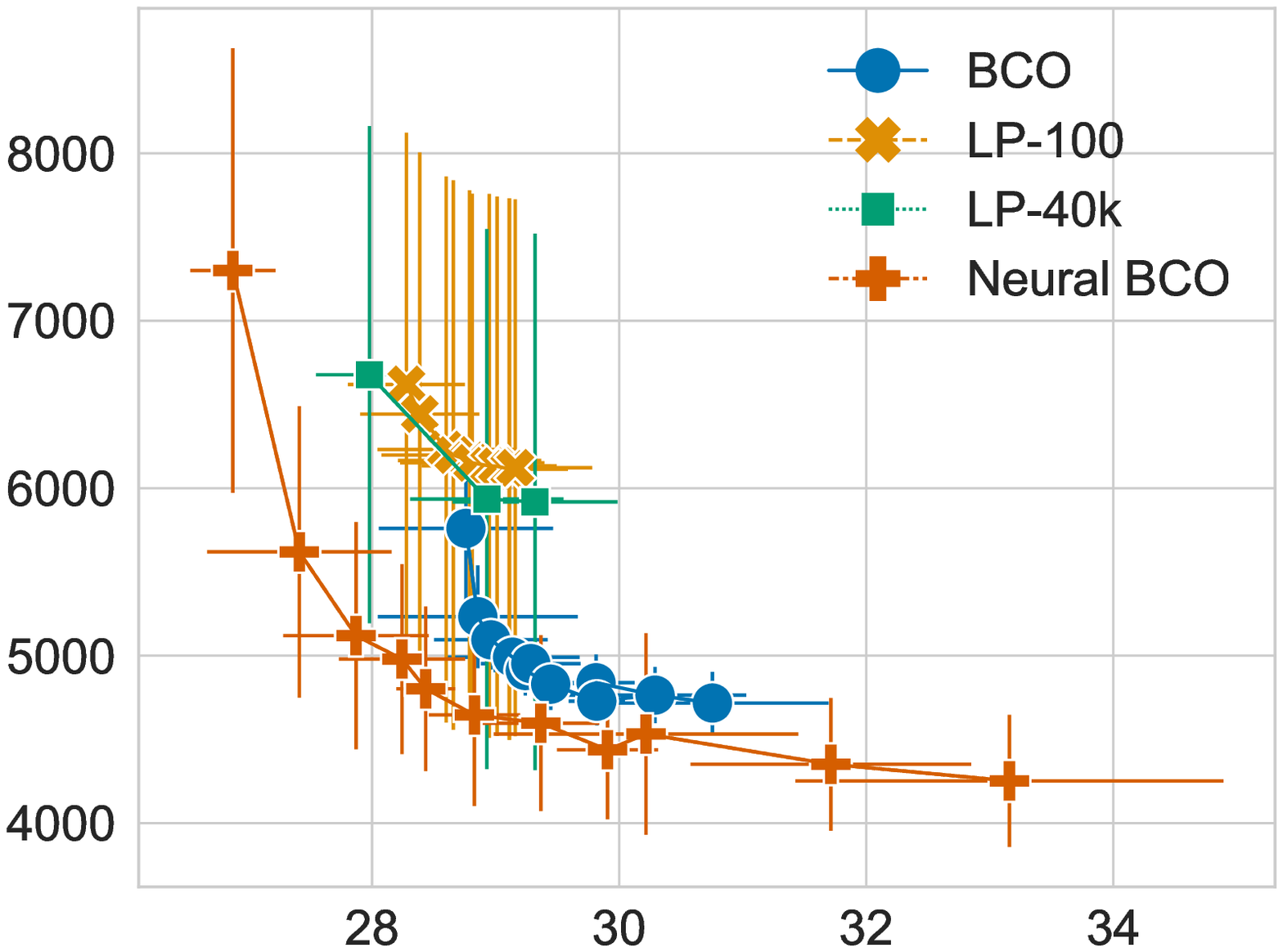}
        \caption{Mumford2}
        \label{subfig:mumford2pareto}
    \end{subfigure}
    \hfill
    \begin{subfigure}[b]{0.32\textwidth}
        \centering
        \includegraphics[width=\textwidth]{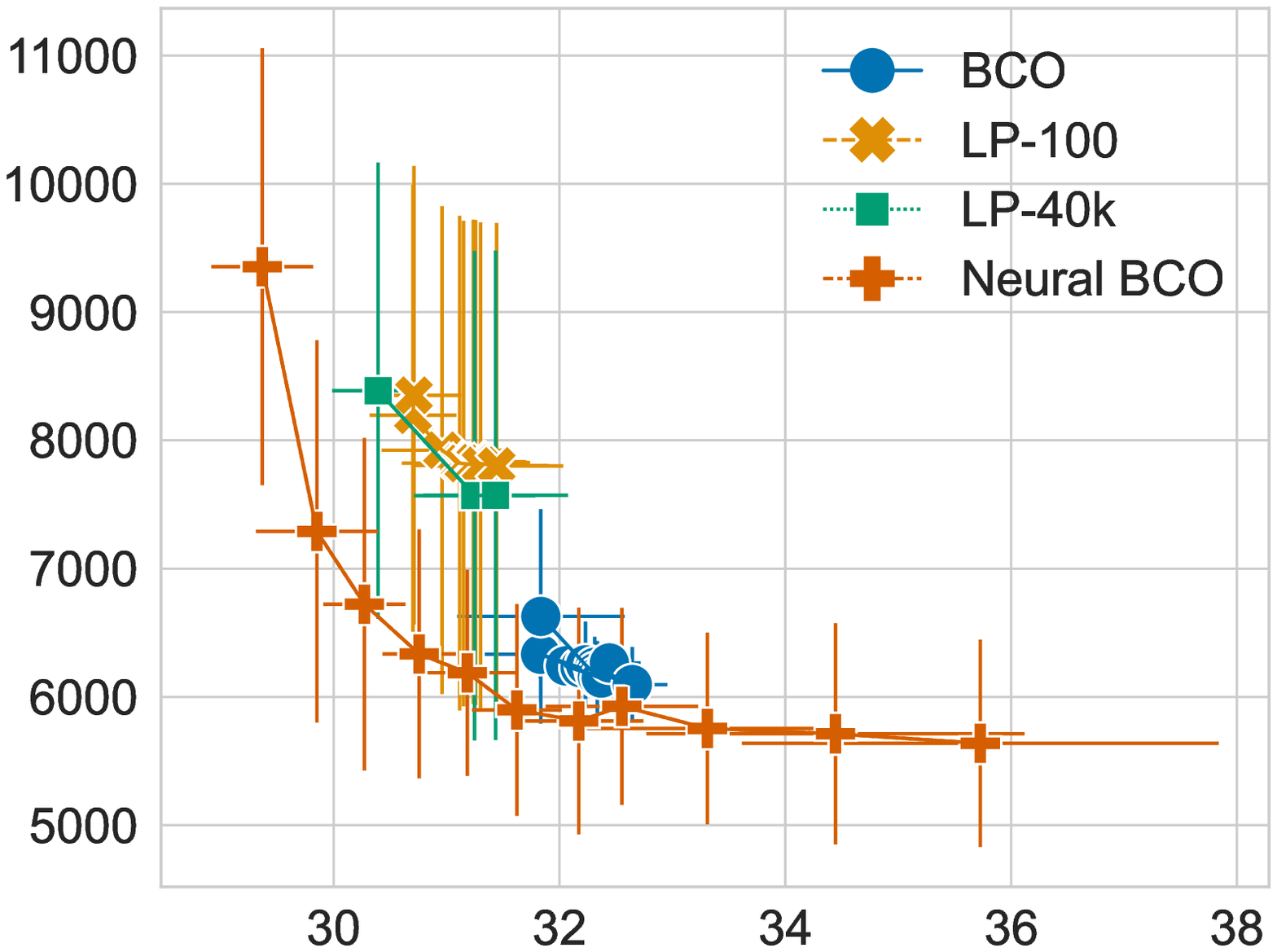}
        \caption{Mumford3}
        \label{subfig:mumford3pareto}
    \end{subfigure}
    \hfill
    \begin{subfigure}[b]{0.32\textwidth}
        \centering
        \includegraphics[width=\textwidth]{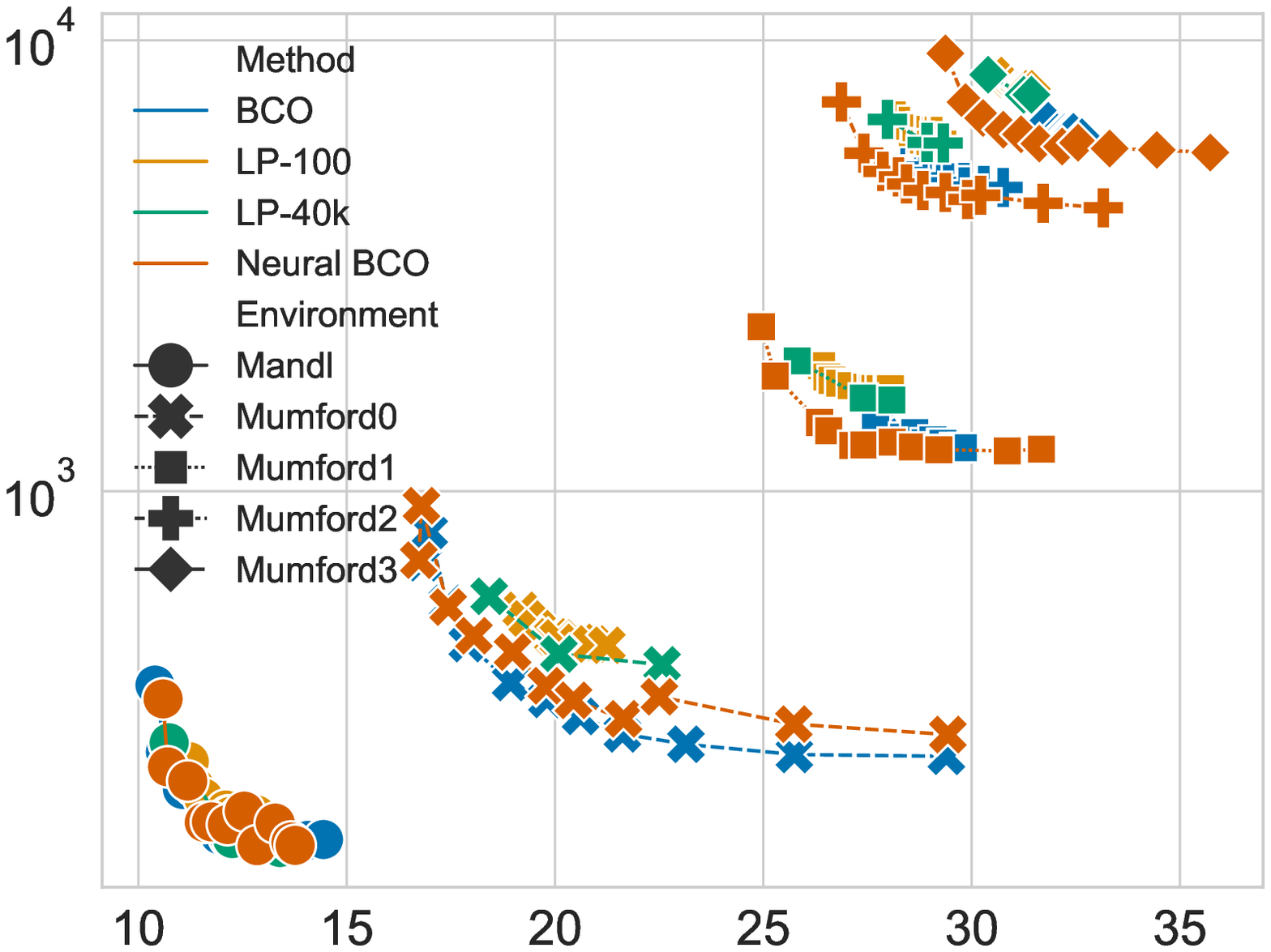}
        \caption{All cities}
        \label{subfig:allpareto}
    \end{subfigure}    
    \caption{Trade-offs achieved by different methods between passenger cost $C_p$ (on the x-axis) and operator cost $C_o$ (on the y-axis), across values of $\alpha$ evenly spaced over the range $[0, 1]$, averaged over 10 random seeds.  Both axes are in units of minutes.  We wish to minimize both values, so the lower-left direction in each plot represents improvement.  The y-axis of sub-figure~\ref{subfig:allpareto} is log-scaled to better fit all curves without flattening them, while the rest are linear.  A line links two points if they have adjacent $\alpha$ values, so these curves show a smooth progression from low $C_o$ to low $C_p$ as $\alpha$ increases.}
    \label{fig:pareto}
\end{figure}

There is a necessary trade-off between minimizing the passenger cost $C_p$ and the operator cost $C_o$: making transit routes longer increases $C_o$, but allows more and faster direct connections between stops, and so may decrease $C_p$.  The weight $\alpha$ can be set by the user to indicate how much they care about $C_p$ versus $C_o$, and each algorithm output will change accordingly.  Figure~\ref{fig:pareto} illustrates the trade-offs made by the different methods, as we vary $\alpha$ over its full range $[0, 1]$ in steps of $0.1$, except for LP-40k, for which we only plot values for $\alpha=0.0, 0.5,$ and $1.0$.  For the two smallest cities (sub-figures~\ref{subfig:mandlpareto} and~\ref{subfig:mumford0pareto}), \ac{BCO} offers a superior trade-off for most $\alpha$, but for the larger cities Mumford1, 2, and 3, NBCO's solutions not only dominate those of \ac{BCO}, they achieve a much wider range of $C_p$ and $C_o$ than either of \ac{BCO} or LP, which will be more satisfactory if the user cares only about one or the other component.  

Both LP and \ac{BCO} have more narrow ranges of $C_p$ and $C_o$ on the three larger cities, but the ranges are mostly non-overlapping.  Some of NBCO's greater range seems to be due to combining the non-overlapping ranges of the constituent parts, but NBCO's range is greater than the union of LP's and \ac{BCO}'s ranges.  This implies that the larger action space of the neural bee versus the type-1 bee allows NBCO to explore a much wider range of solutions by taking wider ``steps'' in solution space.  Meanwhile, LP-40k has about the same range as LP-100, implying that these wide steps must be guided by the metaheuristic to eventually reach a wider space of solutions.

\section{Discussion}

\subsection{Limitations}

\acl{BCO} is just one instance of the broad class of metaheuristic algorithms, and while the results we present are promising, it remains to be seen whether incorporating learned heuristics into metaheuristic algorithms is a sound algorithmic strategy in general.  Furthermore, in this work we consider such a combination in light of only one \ac{CO} problem, the \ac{NDP}.  This is a very interesting and impactful problem, but evaluating this method on a wider variety of \ac{CO} problems such as the \ac{TSP} and \ac{VRP} would more broadly establish the usefulness of this strategy.

We also note that, while the Mumford dataset is a widely used benchmark, it is still synthetic data.  Establishing whether our method would be useful for transit planning in real-world cities will require evaluating on a real-world dataset.

\subsection{Conclusions}

Ultimately, it is doubtful whether a single-pass generation heuristic like that implemented by the \ac{GNN} will be capable of outperforming search based methods like metaheuristics on combinatorial optimization problems like the \ac{NDP}.
By these problems' nature, there is no one-step algorithm for finding optimal solutions, and any fast-to-compute heuristic will necessarily be approximate.  Consequently, methods for exploring the solution space on a given instance have a general advantage over such heuristics.  But we have shown that the choice of heuristics can have a significant impact on the quality of the solutions found by a metaheuristic, and learned heuristics in particular can significantly benefit metaheuristic algorithms when used as some of their sub-heuristics.

In terms of the applicability of these methods in real cities, we note that both LP and Neural \ac{BCO} outperform \ac{BCO} on all three cities - Mumford1, 2, and 3 - that were designed to match a specific real-world city in scale.  Furthermore, the gap between \ac{BCO} and the other methods grows with the size of the city.  This suggests that Neural \ac{BCO} may scale better to much larger problem sizes - which is significant, as some real-world cities have hundreds or even thousands of bus stop locations~\citep{montrealTransit}.

We note that better results could likely be achieved by training a policy directly in a metaheuristic context, rather than training it in isolation and then applying it in a metaheuristic as was done here.  It would also be interesting to use multiple separately-trained models as different heuristics within a metaheuristic algorithm, as opposed to the single model used in our experiments.  This could be seen as a form of ensemble method, with the metaheuristic intelligently combining the strengths of the different learned models to get the best use out of each.

We would also like to explore the training of a further \ac{ML} component to act as the higher-level metaheuristic, creating an entirely learned method for searching the solution space for particular problem instances.  Recent work on few-shot adaptation in \ac{RL}~\citep{team2023human} may provide a promising starting point.

Beyond studying the relative performance of machine learning and metaheuristic approaches and their combination, we hope by this work to draw the attention of the machine community to the \ac{NDP}.  It is a uniquely challenging combinatorial optimization problem with real-world impact, with much potential for novel and useful study by our discipline.


\bibliographystyle{plainnat}
\bibliography{references}

\begin{thebibliography}{46}
\providecommand{\natexlab}[1]{#1}
\providecommand{\url}[1]{\texttt{#1}}
\expandafter\ifx\csname urlstyle\endcsname\relax
  \providecommand{\doi}[1]{doi: #1}\else
  \providecommand{\doi}{doi: \begingroup \urlstyle{rm}\Url}\fi

\bibitem[Ahmed et~al.(2019)Ahmed, Mumford, and Kheiri]{ahmed2019hyperheuristic}
Leena Ahmed, Christine Mumford, and Ahmed Kheiri.
\newblock Solving urban transit route design problem using selection
  hyper-heuristics.
\newblock \emph{European Journal of Operational Research}, 274\penalty0
  (2):\penalty0 545--559, 2019.

\bibitem[Ai et~al.(2022)Ai, Zuo, Chen, and Wu]{ai2022deep}
Guanqun Ai, Xingquan Zuo, Gang Chen, and Binglin Wu.
\newblock Deep reinforcement learning based dynamic optimization of bus
  timetable.
\newblock \emph{Applied Soft Computing}, 131:\penalty0 109752, 2022.

\bibitem[Applegate et~al.(2001)Applegate, Bixby, Chvátal, and
  Cook]{concordeTspSolver}
David Applegate, Robert~E. Bixby, Vašek Chvátal, and William~J. Cook.
\newblock Concorde tsp solver.
\newblock https://www.math.uwaterloo.ca/tsp/concorde/index.html, 2001.

\bibitem[Battaglia et~al.(2018)Battaglia, Hamrick, Bapst, Sanchez-Gonzalez,
  Zambaldi, Malinowski, Tacchetti, Raposo, Santoro, Faulkner,
  et~al.]{battaglia2018relational}
Peter~W Battaglia, Jessica~B Hamrick, Victor Bapst, Alvaro Sanchez-Gonzalez,
  Vinicius Zambaldi, Mateusz Malinowski, Andrea Tacchetti, David Raposo, Adam
  Santoro, Ryan Faulkner, et~al.
\newblock Relational inductive biases, deep learning, and graph networks.
\newblock \emph{arXiv preprint arXiv:1806.01261}, 2018.

\bibitem[Behbahani et~al.(2023)Behbahani, Bauer, Baumli, Baveja, Bhoopchand,
  Bradley-Schmieg, Chang, Clay, Collister, et~al.]{team2023human}
Feryal Behbahani, Jakob Bauer, Kate Baumli, Satinder Baveja, Avishkar
  Bhoopchand, Nathalie Bradley-Schmieg, Michael Chang, Natalie Clay, Adrian
  Collister, et~al.
\newblock Human-timescale adaptation in an open-ended task space.
\newblock \emph{arXiv preprint arXiv:2301.07608}, 2023.

\bibitem[Bengio et~al.(2021)Bengio, Lodi, and Prouvost]{bengio2021machine}
Yoshua Bengio, Andrea Lodi, and Antoine Prouvost.
\newblock Machine learning for combinatorial optimization: a methodological
  tour d’horizon.
\newblock \emph{European Journal of Operational Research}, 290\penalty0
  (2):\penalty0 405--421, 2021.

\bibitem[Brody et~al.(2021)Brody, Alon, and Yahav]{gatv2conv}
Shaked Brody, Uri Alon, and Eran Yahav.
\newblock How attentive are graph attention networks?, 2021.
\newblock URL \url{https://arxiv.org/abs/2105.14491}.

\bibitem[Bruna et~al.(2013)Bruna, Zaremba, Szlam, and LeCun]{bruna2013spectral}
Joan Bruna, Wojciech Zaremba, Arthur Szlam, and Yann LeCun.
\newblock Spectral networks and locally connected networks on graphs.
\newblock \emph{arXiv preprint arXiv:1312.6203}, 2013.

\bibitem[Chien et~al.(2002)Chien, Ding, and Wei]{chien2002dynamic}
Steven I-Jy Chien, Yuqing Ding, and Chienhung Wei.
\newblock Dynamic bus arrival time prediction with artificial neural networks.
\newblock \emph{Journal of transportation engineering}, 128\penalty0
  (5):\penalty0 429--438, 2002.

\bibitem[Dai et~al.(2017)Dai, Khalil, Zhang, Dilkina, and
  Song]{dai2017learningCombinatorial}
Hanjun Dai, Elias~B Khalil, Yuyu Zhang, Bistra Dilkina, and Le~Song.
\newblock Learning combinatorial optimization algorithms over graphs.
\newblock \emph{arXiv preprint arXiv:1704.01665}, 2017.

\bibitem[Darwish et~al.(2020)Darwish, Khalil, and
  Badawi]{darwish2020optimising}
Ahmed Darwish, Momen Khalil, and Karim Badawi.
\newblock Optimising public bus transit networks using deep reinforcement
  learning.
\newblock In \emph{2020 IEEE 23rd International Conference on Intelligent
  Transportation Systems (ITSC)}, pages 1--7. IEEE, 2020.

\bibitem[Defferrard et~al.(2016)Defferrard, Bresson, and
  Vandergheynst]{defferrard2016spectral}
Micha{\"{e}}l Defferrard, Xavier Bresson, and Pierre Vandergheynst.
\newblock Convolutional neural networks on graphs with fast localized spectral
  filtering.
\newblock \emph{CoRR}, abs/1606.09375, 2016.
\newblock URL \url{http://arxiv.org/abs/1606.09375}.

\bibitem[Duvenaud et~al.(2015)Duvenaud, Maclaurin, Iparraguirre, Bombarell,
  Hirzel, Aspuru-Guzik, and Adams]{duvenaud2015convolutional}
David~K Duvenaud, Dougal Maclaurin, Jorge Iparraguirre, Rafael Bombarell,
  Timothy Hirzel, Al{\'a}n Aspuru-Guzik, and Ryan~P Adams.
\newblock Convolutional networks on graphs for learning molecular fingerprints.
\newblock \emph{Advances in neural information processing systems}, 28, 2015.

\bibitem[Fortune(1995)]{fortune1995voronoi}
Steven Fortune.
\newblock Voronoi diagrams and delaunay triangulations.
\newblock \emph{Computing in Euclidean geometry}, pages 225--265, 1995.

\bibitem[Gilmer et~al.(2017)Gilmer, Schoenholz, Riley, Vinyals, and
  Dahl]{gilmer2017quantum}
Justin Gilmer, Samuel~S. Schoenholz, Patrick~F. Riley, Oriol Vinyals, and
  George~E. Dahl.
\newblock Neural message passing for quantum chemistry.
\newblock In Doina Precup and Yee~Whye Teh, editors, \emph{Proceedings of the
  34th International Conference on Machine Learning}, volume~70 of
  \emph{Proceedings of Machine Learning Research}, pages 1263--1272. PMLR,
  06--11 Aug 2017.
\newblock URL \url{https://proceedings.mlr.press/v70/gilmer17a.html}.

\bibitem[Guan et~al.(2006)Guan, Yang, and
  Wirasinghe]{guan2006AnalyticRoutePlanning}
J.F. Guan, Hai Yang, and S.C. Wirasinghe.
\newblock Simultaneous optimization of transit line configuration and passenger
  line assignment.
\newblock \emph{Transportation Research Part B: Methodological}, 40:\penalty0
  885--902, 12 2006.
\newblock \doi{10.1016/j.trb.2005.12.003}.

\bibitem[Guihaire and Hao(2008)]{guihaire2008transitReview}
Val{\'e}rie Guihaire and Jin-Kao Hao.
\newblock Transit network design and scheduling: A global review.
\newblock \emph{Transportation Research Part A: Policy and Practice},
  42\penalty0 (10):\penalty0 1251--1273, 2008.

\bibitem[Jeong and Rilett(2004)]{jeong2004bus}
Ranhee Jeong and R~Rilett.
\newblock Bus arrival time prediction using artificial neural network model.
\newblock In \emph{Proceedings. The 7th international IEEE conference on
  intelligent transportation systems (IEEE Cat. No. 04TH8749)}, pages 988--993.
  IEEE, 2004.

\bibitem[Jiang et~al.(2018)Jiang, Fan, Liu, Zhu, and
  Gu]{jiang2018passengerInflow}
Zhibin Jiang, Wei Fan, Wei Liu, Bingqin Zhu, and Jinjing Gu.
\newblock Reinforcement learning approach for coordinated passenger inflow
  control of urban rail transit in peak hours.
\newblock \emph{Transportation Research Part C: Emerging Technologies},
  88:\penalty0 1--16, 2018.
\newblock ISSN 0968-090X.
\newblock \doi{https://doi.org/10.1016/j.trc.2018.01.008}.
\newblock URL
  \url{https://www.sciencedirect.com/science/article/pii/S0968090X18300111}.

\bibitem[John et~al.(2014)John, Mumford, and Lewis]{john2014routing}
Matthew~P. John, Christine~L. Mumford, and Rhyd Lewis.
\newblock An improved multi-objective algorithm for the urban transit routing
  problem.
\newblock In Christian Blum and Gabriela Ochoa, editors, \emph{Evolutionary
  Computation in Combinatorial Optimisation}, pages 49--60, Berlin, Heidelberg,
  2014. Springer Berlin Heidelberg.
\newblock ISBN 978-3-662-44320-0.

\bibitem[Kepaptsoglou and Karlaftis(2009)]{kepaptsoglou2009transitReview}
Konstantinos Kepaptsoglou and Matthew Karlaftis.
\newblock Transit route network design problem: Review.
\newblock \emph{Journal of Transportation Engineering}, 135\penalty0
  (8):\penalty0 491--505, 2009.
\newblock \doi{10.1061/(ASCE)0733-947X(2009)135:8(491)}.

\bibitem[Kipf and Welling(2016)]{kipf2016semi}
Thomas~N Kipf and Max Welling.
\newblock Semi-supervised classification with graph convolutional networks.
\newblock \emph{arXiv preprint arXiv:1609.02907}, 2016.

\bibitem[Kool et~al.(2019)Kool, Hoof, and Welling]{Kool2019AttentionLT}
W.~Kool, H.~V. Hoof, and M.~Welling.
\newblock Attention, learn to solve routing problems!
\newblock In \emph{ICLR}, 2019.

\bibitem[Kılıç and Gök(2014)]{kilic2014demand}
Fatih Kılıç and Mustafa Gök.
\newblock A demand based route generation algorithm for public transit network
  design.
\newblock \emph{Computers \& Operations Research}, 51:\penalty0 21--29, 2014.
\newblock ISSN 0305-0548.
\newblock \doi{https://doi.org/10.1016/j.cor.2014.05.001}.
\newblock URL
  \url{https://www.sciencedirect.com/science/article/pii/S0305054814001300}.

\bibitem[Li et~al.(2020)Li, Bai, Liu, Yao, and Waller]{li2020graph}
Can Li, Lei Bai, Wei Liu, Lina Yao, and S~Travis Waller.
\newblock Graph neural network for robust public transit demand prediction.
\newblock \emph{IEEE Transactions on Intelligent Transportation Systems}, 2020.

\bibitem[Lu et~al.(2019)Lu, Zhang, and Yang]{lu2019learning}
Hao Lu, Xingwen Zhang, and Shuang Yang.
\newblock A learning-based iterative method for solving vehicle routing
  problems.
\newblock In \emph{International Conference on Learning Representations}, 2019.

\bibitem[Mandl(1980)]{mandl1980evaluation}
Christoph~E Mandl.
\newblock Evaluation and optimization of urban public transportation networks.
\newblock \emph{European Journal of Operational Research}, 5\penalty0
  (6):\penalty0 396--404, 1980.

\bibitem[Mirhoseini et~al.(2021)Mirhoseini, Goldie, Yazgan, Jiang, Songhori,
  Wang, Lee, Johnson, Pathak, Nazi, et~al.]{mirhoseini2021graph}
Azalia Mirhoseini, Anna Goldie, Mustafa Yazgan, Joe~Wenjie Jiang, Ebrahim
  Songhori, Shen Wang, Young-Joon Lee, Eric Johnson, Omkar Pathak, Azade Nazi,
  et~al.
\newblock A graph placement methodology for fast chip design.
\newblock \emph{Nature}, 594\penalty0 (7862):\penalty0 207--212, 2021.

\bibitem[Mumford(2013{\natexlab{a}})]{mumford2013dataset}
Christine~L Mumford.
\newblock Download link to the mumford dataset.
\newblock
  \url{https://users.cs.cf.ac.uk/C.L.Mumford/Research\%20Topics/UTRP/CEC2013Supp.zip},
  2013{\natexlab{a}}.
\newblock Accessed: 2023-03-24.

\bibitem[Mumford(2013{\natexlab{b}})]{mumford2013new}
Christine~L Mumford.
\newblock New heuristic and evolutionary operators for the multi-objective
  urban transit routing problem.
\newblock In \emph{2013 IEEE congress on evolutionary computation}, pages
  939--946. IEEE, 2013{\natexlab{b}}.

\bibitem[Nikoli{\'c} and Teodorovi{\'c}(2013)]{nikolic2013transit}
Milo{\v{s}} Nikoli{\'c} and Du{\v{s}}an Teodorovi{\'c}.
\newblock Transit network design by bee colony optimization.
\newblock \emph{Expert Systems with Applications}, 40\penalty0 (15):\penalty0
  5945--5955, 2013.

\bibitem[Quak(2003)]{quak2003bus}
CB~Quak.
\newblock Bus line planning.
\newblock \emph{A passenger-oriented approach of the construction of a global
  line network and an efficient timetable. Master's thesis, Delft University,
  Delft, Netherlands}, 2003.

\bibitem[Rodrigue(1997)]{rodrigueNNsForLandUseAndTransport}
Jean-Paul Rodrigue.
\newblock Parallel modelling and neural networks: An overview for
  transportation/land use systems.
\newblock \emph{Transportation Research Part C: Emerging Technologies},
  5\penalty0 (5):\penalty0 259--271, 1997.
\newblock ISSN 0968-090X.
\newblock \doi{https://doi.org/10.1016/S0968-090X(97)00014-4}.
\newblock URL
  \url{https://www.sciencedirect.com/science/article/pii/S0968090X97000144}.

\bibitem[{Soci\'et\'e de transport de Montr\'eal}(2013)]{montrealTransit}
{Soci\'et\'e de transport de Montr\'eal}.
\newblock Everything about the stm, 2013.
\newblock URL
  \url{https://web.archive.org/web/20130610123159/http://www.stm.info/english/en-bref/a-toutsurlaSTM.htm}.
\newblock Accessed: 2023-05-17.

\bibitem[S{\"o}rensen et~al.(2018)S{\"o}rensen, Sevaux, and
  Glover]{sorensen2018history}
Kenneth S{\"o}rensen, Marc Sevaux, and Fred Glover.
\newblock A history of metaheuristics.
\newblock In \emph{Handbook of heuristics}, pages 791--808. Springer, 2018.

\bibitem[Sykora et~al.(2020)Sykora, Ren, and Urtasun]{sykora2020multi}
Quinlan Sykora, Mengye Ren, and Raquel Urtasun.
\newblock Multi-agent routing value iteration network.
\newblock In \emph{International Conference on Machine Learning}, pages
  9300--9310. PMLR, 2020.

\bibitem[van Nes(2003)]{vannes2003AnalyticRouteAndSchedule}
Rob van Nes.
\newblock Multiuser-class urban transit network design.
\newblock \emph{Transportation Research Record}, 1835\penalty0 (1):\penalty0
  25--33, 2003.
\newblock \doi{10.3141/1835-04}.
\newblock URL \url{https://doi.org/10.3141/1835-04}.

\bibitem[Vaswani et~al.(2017)Vaswani, Shazeer, Parmar, Uszkoreit, Jones, Gomez,
  Kaiser, and Polosukhin]{vaswani2017attention}
Ashish Vaswani, Noam Shazeer, Niki Parmar, Jakob Uszkoreit, Llion Jones,
  Aidan~N Gomez, {\L}ukasz Kaiser, and Illia Polosukhin.
\newblock Attention is all you need.
\newblock In \emph{Advances in neural information processing systems}, pages
  5998--6008, 2017.

\bibitem[Vinyals et~al.(2015)Vinyals, Fortunato, and
  Jaitly]{vinyals2015pointer}
Oriol Vinyals, Meire Fortunato, and Navdeep Jaitly.
\newblock Pointer networks.
\newblock \emph{arXiv preprint arXiv:1506.03134}, 2015.

\bibitem[Williams(1992)]{williams1992reinforce}
Ronald~J Williams.
\newblock Simple statistical gradient-following algorithms for connectionist
  reinforcement learning.
\newblock \emph{Machine learning}, 8\penalty0 (3):\penalty0 229--256, 1992.

\bibitem[Xiong and Schneider(1992)]{xiong1992transportation}
Yihua Xiong and Jerry~B Schneider.
\newblock Transportation network design using a cumulative genetic algorithm
  and neural network.
\newblock \emph{Transportation Research Record}, 1364, 1992.

\bibitem[Yan et~al.(2023)Yan, Cui, Chen, and Ma]{Yan2023DistributedMD}
Haoyang Yan, Zhiyong Cui, Xinqiang Chen, and Xiaolei Ma.
\newblock Distributed multiagent deep reinforcement learning for multiline
  dynamic bus timetable optimization.
\newblock \emph{IEEE Transactions on Industrial Informatics}, 19:\penalty0
  469--479, 2023.

\bibitem[Ying et~al.(2018)Ying, He, Chen, Eksombatchai, Hamilton, and
  Leskovec]{ying2018webscale}
Rex Ying, Ruining He, Kaifeng Chen, Pong Eksombatchai, William~L. Hamilton, and
  Jure Leskovec.
\newblock Graph convolutional neural networks for web-scale recommender
  systems.
\newblock \emph{CoRR}, abs/1806.01973, 2018.
\newblock URL \url{http://arxiv.org/abs/1806.01973}.

\bibitem[Yoo et~al.(2023)Yoo, Lee, and Han]{yoo2023reinforcement}
Sunhyung Yoo, Jinwoo~Brian Lee, and Hoon Han.
\newblock A reinforcement learning approach for bus network design and
  frequency setting optimisation.
\newblock \emph{Public Transport}, pages 1--32, 2023.

\bibitem[Zou et~al.(2006)Zou, Xu, and Zhu]{zou2006lightrail}
Liang Zou, Jian-min Xu, and Ling-xiang Zhu.
\newblock Light rail intelligent dispatching system based on reinforcement
  learning.
\newblock In \emph{2006 International Conference on Machine Learning and
  Cybernetics}, pages 2493--2496, 2006.
\newblock \doi{10.1109/ICMLC.2006.258785}.

\bibitem[Çodur and Tortum(2009)]{akgungorNNsForAccidentPrediction}
Muhammed~Yasin Çodur and Ahmet Tortum.
\newblock An artificial intelligent approach to traffic accident estimation:
  Model development and application.
\newblock \emph{Transport}, 24\penalty0 (2):\penalty0 135--142, 2009.
\newblock \doi{10.3846/1648-4142.2009.24.135-142}.

\end{thebibliography}

\begin{acronym}
  \acro{AV}{autonomous vehicle}
  \acro{TSP}{Travelling Salesman problem}
  \acro{VRP}{Vehicle Routing Problem}  
  \acro{NDP}{Transit Network Design Problem}
  \acro{FSP}{Frequency-Setting Problem}
  \acro{DFSP}{Design and Frequency-Setting Problem}
  \acro{SP}{Scheduling Problem}
  \acro{TP}{Timetabling Problem}
  \acro{NDSP}{Network Design and Scheduling Problem}
  \acro{BCO}{Bee Colony Optimization}
  \acro{HH}{hyperheuristic}
  \acro{GA}{Genetic Algorithm}
  \acro{SA}{Simulated Annealing}
  
  \acro{MoD}{Mobility on Demand}
  \acro{AMoD}{Autonomous Mobility on Demand}
  \acro{IMoDP}{Intermodal Mobility-on-Demand Problem}
  \acro{OD}{Origin-Destination}
  \acro{CSA}{Connection Scan Algorithm}

  \acro{CO}{combinatorial optimization}
  \acro{NN}{neural network}
  \acro{ML}{Machine Learning}
  \acro{MLP}{Multi-Layer Perceptron}
  \acro{RL}{Reinforcement Learning}
  \acro{DRL}{Deep Reinforcement Learning}
  \acro{GNN}{Graph Neural Network}
  \acro{MDP}{Markov Decision Process}
  \acro{DQN}{Deep Q-Networks}
  \acro{ACER}{Actor-Critic with Experience Replay}
  \acro{PPO}{Proximal Policy Optimization}
  \acro{ARTM}{Metropolitan Regional Transportation Authority}
\end{acronym}

\end{document}